%% file: neurips_2026.tex
\documentclass{article}

\usepackage[sglblindworkshop, final]{neurips_2026}
\workshoptitle{The 10th Machine Learning for Systems Workshop at Neurips 2026, Sydney, Australia}

\usepackage[utf8]{inputenc} 
\usepackage[T1]{fontenc}    
\usepackage{hyperref}       
\usepackage{url}            
\usepackage{booktabs}       
\usepackage{amsfonts}       
\usepackage{nicefrac}       
\usepackage{microtype}      
\usepackage{xcolor}         
\usepackage{xspace}
\usepackage{graphicx}
\usepackage{subcaption}
\usepackage{float}

\title{Aligning Training and Rollout Policies for End-to-End FP4 RL on Mixture-of-Experts LLMs}

\author{%
  Hei Yi Mak$^*$, Hoang Le$^*$,  Mehran Taghian Jazi, Yunke Peng, Huan Lin, Zhengzhi Tang, \\
  \textbf{Zhexu Liu, Yao Wang, Yaoyuan Wang, Anandharaju Durai Raju, Tanzila Rahman}  \\
  Huawei  \\
  \texttt{\{hei.yi.mak1, pengyunke\}@huawei.com} \\
}

\begin{document}
\newcommand{\ourmethod}{MoEResQ\xspace}
\newcommand{\comment}[1]{{\color{blue} \textbf{Comment:} #1}}

\maketitle
{%
  \renewcommand{\thefootnote}{\fnsymbol{footnote}}%
  \footnotetext[1]{Equal contribution.}%
}

\begin{abstract}
  In this work, we study end-to-end FP4 reinforcement learning (RL) post-training for Mixture-of-Experts (MoE) large language models. Recent work on FP4 RL has introduced effective recovery recipes that restore accuracy under aggressive low-precision quantization for dense models, but we find that the same approach yields no improvement over the naive FP4 baseline in the MoE setting. We conduct an in-depth study to understand why. Our analysis identifies the root cause in the routing mechanism unique to MoE: because the router makes discrete expert-selection decisions, training–inference policy alignment becomes acutely sensitive to whether the same recovery technique is applied symmetrically on the training and inference sides. When an asymmetric recipe is used—including the dense-model approach, which corrects only the rollout—training–inference alignment deteriorates sharply, as measured by the Pearson correlation coefficient between training and inference probabilities, and router selection matching rates. Building on this finding, we propose \ourmethod, a quantization-recovery recipe for MoE RL post-training that enforces symmetry between the training- and inference-side corrections. \ourmethod restores training–inference alignment and substantially closes the accuracy gap to full precision, establishing symmetric recovery as the key principle in FP4 RL for MoE models.
\end{abstract}

\input{sections/introduction}
\input{sections/background}
\input{sections/alignment}

\input{sections/method}
\input{sections/experiment}
\input{sections/conclusion}

\bibliographystyle{unsrt}
\bibliography{neurips_2026}







\end{document}

%% file: sections/introduction.tex
\section{Introduction}
Large Language Models (LLMs) have achieved strong performance across a wide range of tasks, including mathematical problem solving, code generation, and agentic workflows \cite{brown2020language, ouyang2022training}. Reinforcement Learning with Verifiable Rewards, using policy-gradient algorithms such as GRPO~\cite{shao2024deepseekmath}, has become a successful paradigm for LLM post-training \cite{deepseek2017r1}. Despite its effectiveness for reasoning, RL post-training is highly resource-intensive in both memory and computation. Quantization has emerged as an effective means of reducing memory footprint and computational cost. 
Prior work has explored low-precision computation for RL, with 8-bit formats on the rollout side, and \cite{qurl} on both the training and rollout policies \cite{jetrl}. Most recently, QeRL\cite{huang2026qerl} applied weight-only quantization to the rollout policy in combination with LoRA~\cite{hu2021lora}, keeping activations and gradients in higher precision, while RolloutResQ~\cite{rolloutresq} introduced a residual correction for the rollout activations and quantized both the training and rollout sides to 4-bit. A central challenge shared by these methods is the mismatch between the training and rollout sides, which has so far been studied only for dense models~\cite{yao2025rollout}.

Mixture-of-Experts (MoE) LLMs, which combine dynamic routing with sparse activation, attain higher performance at reduced computational  \cite{liu2024deepseek, jiang2024mixtral}, making them a natural next target for low-precision RL. In this paper, we ask whether RolloutResQ, the quantization recovery recipe developed for dense models in FP4 RL, transfers to MoE models.
Through a systematic study, we find that 
RolloutResQ 
fails to recover any accuracy over the FP4 baseline in the MoE setting.
We trace this failure to the training--rollout mismatch introduced by the routing mechanism, whose expert selection is highly sensitive to quantization error. In particular, an \emph{asymmetric} recipe---in which the quantization recovery method adopted in training differs from that used in rollout---suffers substantially more degradation than a \emph{symmetric} recipe that applies the same method on both sides.

Guided by this finding, we propose \ourmethod, which adapts RolloutResQ to align the training and rollout sides by applying a residual correction on the training side as well. We further study a fine-grained configuration that quantizes the linear layers of the experts only. Together, these choices close the training-inference alignment as well as the accuracy gap to full precision (BF16) for FP4 RL on MoE models.

\input{tables/eval_results}

%% file: tables/eval_results.tex
\begin{table}[htbp]
\centering
\footnotesize
\begin{tabular}{lcc|cc|cc}
\toprule
\textbf{Dataset} & 
\textbf{BF16} & 
\textbf{HiF4} & 
\textbf{RolloutResQ} & 
\textbf{\ourmethod{}} &
\textbf{\begin{tabular}[t]{@{}c@{}}RolloutResQ\\(experts only)\end{tabular}} & 
\textbf{\begin{tabular}[t]{@{}c@{}}\ourmethod{}\\(experts only)\end{tabular}} \\
\midrule
\midrule
\textbf{AMC23} & 85.10 & 65.23 \small{(-19.87)} & 62.34 \small{(-22.76)}  & \textbf{84.45} \small{(-0.65)} & 78.75 \small{(-6.35)}& \textbf{84.61 \small{(-0.49)}} \\
\textbf{AIME24} & 48.23 & 23.96 \small{(-24.27)} & 25.00 \small{(-23.23)} & \textbf{44.48} \small{(-3.75)} & 37.39 \small{(-10.84)}& \textbf{46.98 \small{(-1.25)}} \\
\textbf{AIME25} & 34.38 & 20.83 \small{(-13.55)} & 19.06 \small{(-15.32)}  & \textbf{29.69} \small{(-4.69)} & 29.06 \small{(-5.32)} & \textbf{32.92 \small{(-1.46)}} \\
\textbf{MATH500} & 86.71 & 79.21 \small{(-7.50)} & 74.00 \small{(-12.71)} & \textbf{85.42} \small{(-1.29)} & 82.46 \small{(-4.25)} & \textbf{86.17 \small{(-0.54)}} \\
\bottomrule
\end{tabular}%
\vspace{4pt}
\caption{Evaluation results for Qwen3-30B-A3B trained on DAPO-Math-17K with GRPO. For HiF4, RolloutResQ and \ourmethod, the attention projections and expert projections are in HiF4 precision. For RolloutResQ (experts only) and \ourmethod (experts only), only expert projections are in HiF4 precision. In all experiments, the routing layers, constituting only 0.041\% of the total parameters, are kept in BF16 as outputs are highly sensitive to routing selection.}
\label{tab:eval_results}
\vspace{-20pt}
\end{table}

%% file: sections/background.tex
\section{Preliminaries}
\subsection{Residual Correction and RolloutResQ} Prior work has identified activation outliers as a primary obstacle to low-precision quantization. As highlighted in \cite{rolloutresq, zhang2025accurate, wang2025optimizing}, such outliers widen the dynamic range of the represented values, forcing the limited FP4 grid to allocate its few available levels across an inflated range and thereby collapsing a large fraction of smaller values to zero. \cite{rolloutresq, zhang2025accurate} address this issue by \emph{residual correction} (RC), which computes an activation residual and uses it to correct the output. Formally, for each linear layer $\ell$, the activation residual is given by $\Delta X_\ell = X_\ell - Q(X_\ell)$, where $Q$ is the quantization operator. The residual is then quantized to recover the precision loss through two FP4 GEMMs,
\begin{equation}
\label{eq:of}
\hat{Y_\ell}
=
Q(X_\ell)Q(W_\ell)^\top
+
Q(\Delta X_\ell)Q(W_\ell)^\top.
\end{equation}
In FP4 RL post-training setting, \cite{rolloutresq} identifies rollout activation is the primary source of degradation and propose an effective recipe, RolloutResQ, to recover accuracy loss for dense LLMs, which applies residual correction on rollout activations. In this work, we use HiF4~\cite{luo2026hifloat4formatlanguagemodel} as the FP4 quantization format and do not quantize the routers of the MoE models as they constitute only $0.38\% $ of the active parameters and the outputs are sensitive to routing selection. Due to hardware limitations, all experiments are conducted under simulated quantization.

\subsection{Mixture-of-Experts LLMs}
Mixture-of-Experts (MoE) LLMs have recently emerged as a leading recipe for scaling
model capacity without a proportional increase in computational cost. By replacing the
dense feed-forward networks (FFNs) of a standard Transformer with a collection of expert
FFNs and a lightweight router that activates only the top-$k$ experts per token, MoE models
decouple total parameter count from per-token compute~\cite{shazeer2017outrageously,fedus2022switch}.
This design has powered some of the strongest open-weight reasoning models to date:
DeepSeek-V4~\cite{deepseekai2026deepseekv4highlyefficientmilliontoken}, Qwen3~\cite{yang2025qwen3}, Kimi-K3~\cite{kimiteam2026kimik3openfrontier}.
Given this combination of
strong reasoning ability and favorable compute--performance trade-off, MoE models are a natural
target for low-precision RL post-training.

This sparsity, however, comes at the cost of training difficulty. Because the router performs a
\emph{discrete}, input-dependent top-$k$ expert selection, MoE training is prone to instabilities
such as \emph{routing collapse}~\cite{shazeer2017outrageously,fedus2022switch, chen2025moequant}, and is unusually
sensitive to small perturbations of the router's input activations. This sensitivity is amplified
in low-precision RL: when rollout and training run under different numerical conditions, the router
can select \emph{different} experts for the same token, so a response is generated under one
effective policy but scored under another---a rollout--training mismatch that widens the off-policy
gap and can severely degrade or even collapse training~\cite{ma2025rolloutroutingreplay}. This same
sensitivity is what makes the interaction between MoE routing and quantization non-trivial, and it
motivates the analysis in the following sections.

%% file: sections/alignment.tex
\input{tables/visualization}
\input{tables/alignment}

\section{Symmetry Restores Training–Inference Alignment in FP4 MoE RL}

Table \ref{tab:eval_results} contains the evaluation results of training Qwen3-30B-A3B \cite{yang2025qwen3} with GRPO in BF16 precision and HiF4 precision. The degradation is significant, with 19.87\%, 24.27\%, 13.55\%, 7.50\% accuracy drops across the evaluation AMC-2023~\cite{amc23}, AIME-2024~\cite{jia2024aime2024}, AIME-2025~\cite{aime2025_hf}, and MATH-500~\cite{hendrycks2021measuring}, respectively.
While RolloutResQ shows great success in recovering accuracy on dense models~\cite{rolloutresq}; we ask whether it transfers to the MoE architecture. 
The outcome is not merely a smaller gain—it is a reversal. As shown in Table~\ref{tab:eval_results},
applying RolloutResQ fails to recover accuracy and, on three of four benchmarks, performs
even \emph{worse} than the naive HiF4 baseline that applies no recovery at all. On AMC23 it drops
$22.76$ points from BF16 ($62.34$ vs.\ $85.10$), below HiF4's already-degraded $65.23$;
on AIME25 it falls to $19.06$, below HiF4's $20.83$; and on MATH500 it reaches
only $74.00$, below HiF4's $79.21$. 

The disappointing outcome naturally leads to a key question: Why does a low precision RL recipe that works on dense models fail so decisively on MoE? We hypothesize that this is due to the asymmetric nature of RolloutResQ, where residual correction is applied only on inference side but not on training side. In the MoE setting this mismatch can be amplified by a mechanism absent in dense models: the router. Because expert selection is a discrete argmax over gate logits, a small perturbation to activations does not merely shift an output—it can flip which experts are executed. When correction is applied during rollout but not during training, this can encourage the inference and training policies to route tokens to different experts, and the resulting divergence compounds. 

Table \ref{tab:model_metrics_transposed} supports this hypothesis. 
It compares the training and inference alignment on rollout generated from the inference policy with vLLM, using three metrics: 
the Pearson correlation coefficient (PCC) between training and inference probabilities, \texttt{top\_k\_full\_match\_frac}, and \texttt{topk\_overlap\_mean}. 
Under the full-scope setting (quantizing experts and attention projections), one-sided RC 
leaves alignment near the HiF4 level: PCC is $0.9444$ (\textbf{RC Inference}) and $0.9580$ (\textbf{RC Training}), against HiF4's $0.9410$ and BF16's $0.9951$;  \texttt{topk\_overlap\_mean} and \texttt{top\_k\_full\_match\_frac} exhibit the same pattern---RC Inference and RC Training hover around HiF4's level and remain far from BF16.
Applying RC to both sides (\textbf{Both}) restores every metric back to near-BF16: PCC $0.9940$,
overlap $0.9840$, 
and \texttt{top\_k\_full\_match\_frac} recovering close to BF16 standard at all five layers. Figure \ref{fig:prob_figures} further illustrates the alignment between training and inference probabilities across configurations. Notably, the points for RC Both cluster much more tightly along the diagonal (perfect alignment) than those for HiF4 and the asymmetric recipes, closely mirroring BF16. The same behavior holds when quantization is restricted to expert layers only (\textbf{experts only}), where the symmetric recipe recovers alignment \emph{almost fully} on all metrics.
In every scope and at every layer, applying RC on only one side brings nearly no improvement over HiF4,
but applying to both sides restores every alignment metric to BF16 level, demonstrating that symmetry is the key to closing the alignment gap.

%% file: tables/visualization.tex
\begin{figure*}[htbp]
    \vspace{-4pt}
    \centering
    \begin{subfigure}[b]{0.18\textwidth}
        \centering
        \includegraphics[width=\textwidth]{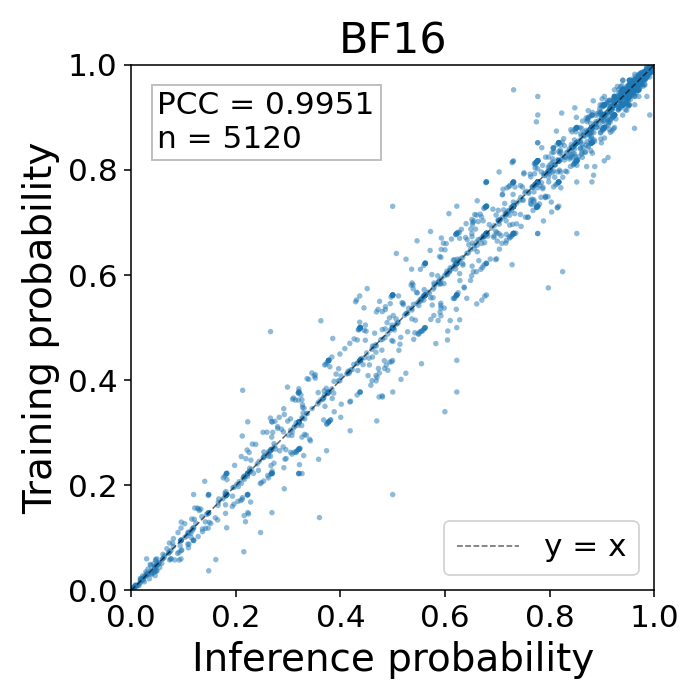}
        \label{fig:method_a}
    \end{subfigure}
    \hfill 
    \begin{subfigure}[b]{0.18\textwidth}
        \centering
        \includegraphics[width=\textwidth]{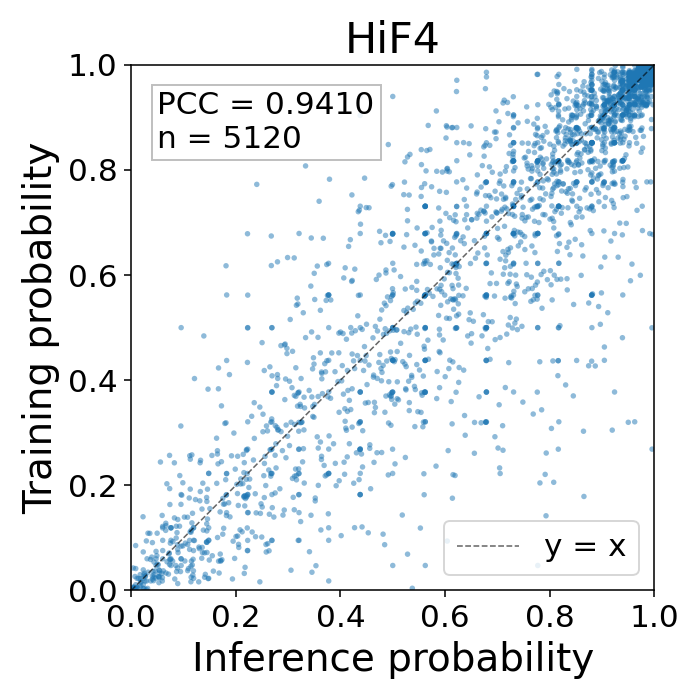}
        \label{fig:method_b}
    \end{subfigure}
    \hfill
    \begin{subfigure}[b]{0.18\textwidth}
        \centering
        \includegraphics[width=\textwidth]{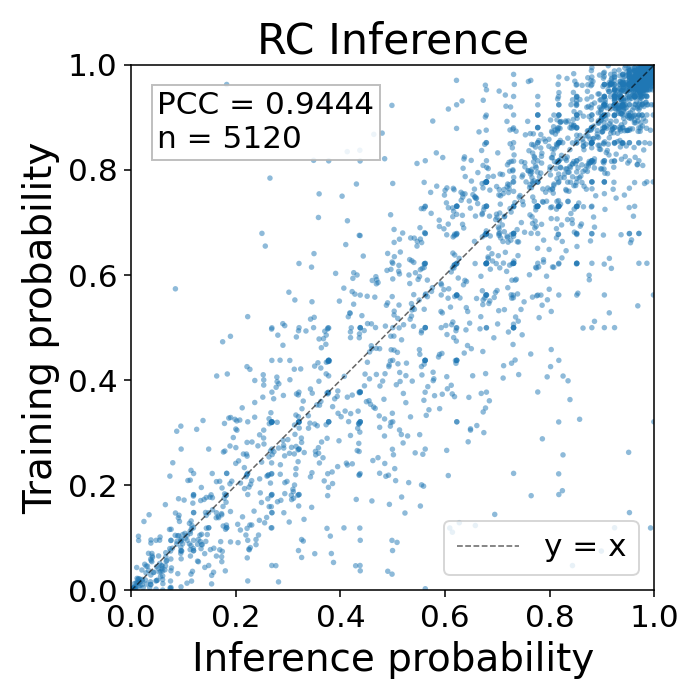}
        \label{fig:method_c}
    \end{subfigure}
    \hfill
    \begin{subfigure}[b]{0.18\textwidth}
        \centering
        \includegraphics[width=\textwidth]{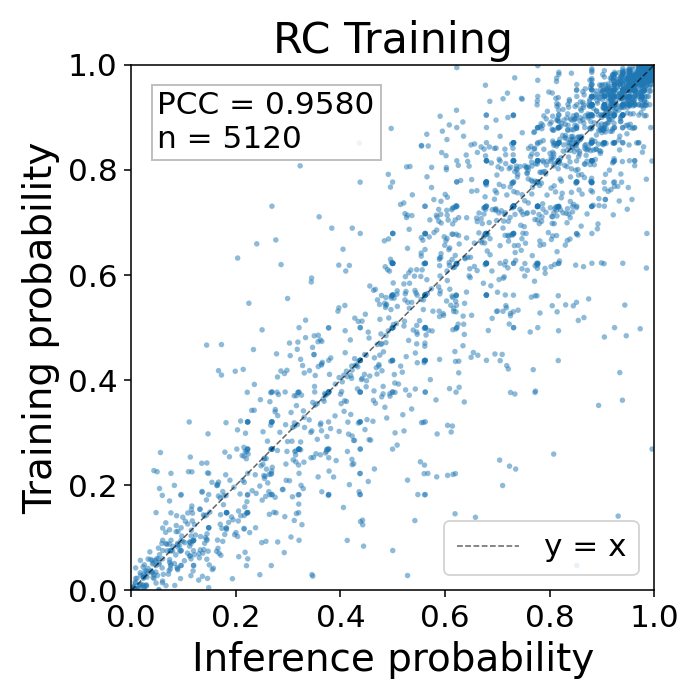}
        \label{fig:method_d}
    \end{subfigure}
    \hfill
    \begin{subfigure}[b]{0.18\textwidth}
        \centering
        \includegraphics[width=\textwidth]{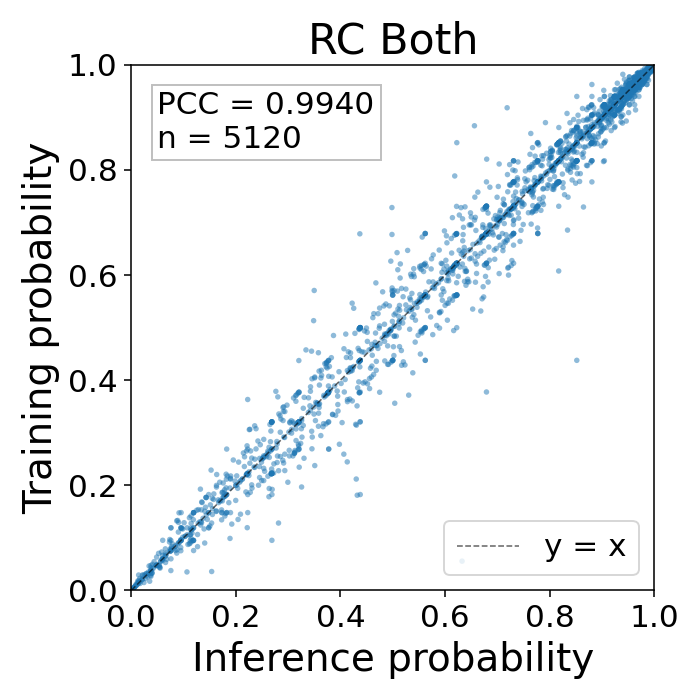}
        \label{fig:method_ours}
    \end{subfigure}
    \vspace{-8pt}
    \caption{Visualization of the alignment between training and inference probabilities for BF16, HiF4, RC Inference, RC Training, and RC Both.}
    \label{fig:prob_figures}
    \vspace{-4pt}
\end{figure*}

%% file: tables/alignment.tex
\begin{table}[htbp]
\centering
\footnotesize
\setlength{\tabcolsep}{3.5pt} 
\begin{tabular}{l c c ccc ccc}
\toprule
 & & & \multicolumn{3}{c}{\textbf{RC}} & \multicolumn{3}{c}{\textbf{RC (Experts Only)}} \\
\cmidrule(lr){4-6} \cmidrule(lr){7-9}
\textbf{Metric} & \textbf{BF16} & \textbf{HiF4} & \textbf{Inference} & \textbf{Training} & \textbf{Both} & \textbf{Inference} & \textbf{Training} & \textbf{Both} \\
\midrule
\textbf{PCC} & 0.9951 & 0.9410 & 0.9444 & 0.9580 & \textbf{0.9940} & 0.9702 & 0.9796 & \textbf{0.9952} \\
\midrule
\multicolumn{9}{l}{\textbf{top\_k\_match\_frac}} \\
\quad Layer 0  & 0.9869 & 0.9801 & 0.8100 & 0.8195 & \textbf{0.9709} & 0.9867 & 0.9875 & \textbf{0.9867} \\
\quad Layer 12 & 0.8799 & 0.6172 & 0.6279 & 0.6590 & \textbf{0.8682} & 0.7373 & 0.7879 & \textbf{0.8730} \\
\quad Layer 23 & 0.8562 & 0.5883 & 0.5990 & 0.6402 & \textbf{0.8418} & 0.7209 & 0.7674 & \textbf{0.8506} \\
\quad Layer 35 & 0.8459 & 0.6256 & 0.6514 & 0.6836 & \textbf{0.8359} & 0.7354 & 0.7947 & \textbf{0.8455} \\
\quad Layer 47 & 0.8680 & 0.6926 & 0.7092 & 0.7297 & \textbf{0.8846} & 0.7748 & 0.8059 & \textbf{0.8687} \\
\midrule
\textbf{topk\_overlap\_mean} & 0.9850 & 0.9580 & 0.9550 & 0.9600 & \textbf{0.9840} & 0.9710 & 0.9770 & \textbf{0.9840} \\
\midrule
\end{tabular}
\vspace{4pt}
\caption{Training-inference alignment base on three metrics: the Pearson correlation coefficient (PCC) between training and inference probabilities, the fraction of tokens whose top-$k$ expert set matches exactly (\texttt{top\_k\_match\_frac}), and the mean pairwise overlap of the top-$k$ sets (\texttt{topk\_overlap\_mean}). For \texttt{top\_k\_match\_frac}, we measured the values for layer 0, 12, 23, 35, and 47. RC denotes applying RC when quantizing both attention linear projections and expert linear projections, and RC (Experts only) denotes applying RC when quantizing expert linear projection only.}
\label{tab:model_metrics_transposed}
\vspace{-16pt}
\end{table}

%% file: sections/method.tex
\section{\ourmethod: Method and Results}

Base on the above analysis, we present \ourmethod, an symmetric recipe for recovering the precision loss induced by the training--inference mismatch in MoE models under fully FP4 RL. 
Having identified that applying quantization recovery to only one side—training or inference—is the principal failure mode,
\ourmethod applies RC to the same layers on both training and inference, maximizing the symmetry between the two policies and thereby minimizing the mismatch.
Concretely, during rollout we apply the activation correction of Equation~\ref{eq:of}. During training, the forward pass uses the same residual correction as the inference side (Equation~\ref{eq:of}). 
For the backward pass, given the input $X_\ell$, the weight $W_\ell$, and the output gradient $dY_\ell$, the input and weight gradients are computed as $dX_\ell
=
dY_\ell\,W_\ell$ and $dW_\ell
=
(dY_\ell)^{\top} X_\ell$, respectively.
To correct the quantization error in the backward pass, we apply the residual correction to both $dY_\ell$ and $X_\ell$, so that the computation retains $dY_\ell$, $\Delta dY_\ell$, $X_\ell$, and $\Delta X_\ell$, where $\Delta X_\ell = X_\ell - Q(X_\ell)$ and $\Delta dY_\ell = dY_\ell - Q(dY_\ell)$.

We also consider the setting in which only the expert linear layers are quantized to FP4---the "experts only" scope studied in Section 3. The expert projections account for the majority of the model's parameters—up to 94.66\% in Qwen3-30B-A3B—so quantizing them alone captures most of the efficiency benefit of FP4. The attention linear layers are kept in BF16, which limits perturbation to the router's input and preserves routing quality.


%% file: sections/experiment.tex
Our experiments are conducted using VeRL~\cite{VeRL}, with vLLM~\cite{kwon2023efficient} for rollout and FSDP~\cite{zhao2023pytorch} for training. We train Qwen3-30B-A3B on DAPO-Math-17k~\cite{yu2026dapo} with GRPO and evaluate it on AMC-2023, AIME-2024, AIME-2025, and MATH-500. As shown in Table~\ref{tab:eval_results}, \ourmethod recovers near-BF16 accuracy across all four benchmarks, substantially narrowing the gap between HiF4 and the BF16 baseline: from $19.87\%$ to $0.65\%$ on AMC23, from $24.27\%$ to $3.75\%$ on AIME24, from $13.55\%$ to $4.69\%$ on AIME25, and from $7.5\%$ to $1.29\%$ on MATH500. 
When quantization is applied only to the expert layers, \ourmethod pushes performance even closer to BF16, closing the gap to just $0.49\%$, $1.25\%$, $1.46\%$, and $0.54\%$ on AMC23, AIME24, AIME25, and MATH500, respectively. Remarkably, in this configuration \ourmethod lands within about $0.5\%$ of BF16 on both AMC23 and MATH500, and within $1.5\%$ on all four benchmarks---effectively matching full-precision accuracy in a fully FP4 setting. These results demonstrate that \ourmethod recovers near-BF16 accuracy in end-to-end FP4 RL post-training of MoE LLMs, a low-precision regime in which RolloutResQ is insufficient.

%% file: sections/conclusion.tex
\section{Conclusion}
We studied end-to-end FP4 RL post-training for MoE LLMs and found that RolloutResQ, the dense-model residual-correction recipe, does not transfer to this setting. The cause is the router: since expert selection is a discrete argmax over gate logits, the mismatch from an asymmetric recovery recipe is amplified into divergent routing that degrades training–inference alignment. Our recipe, \ourmethod{}, applies residual correction symmetrically on both sides, restoring alignment and recovering near-BF16 accuracy—revealing that training–inference symmetry is the key to accuracy recovery in FP4 RL for MoE models. 
A promising direction for future work is to push the efficiency of this recipe further while sustaining near-BF16 accuracy.

